\documentclass{article}

\usepackage[final]{neurips_2023}

\usepackage[utf8]{inputenc} 
\usepackage[T1]{fontenc}    
\usepackage{hyperref}       
\usepackage{url}            
\usepackage{booktabs}       
\usepackage{amsfonts}       
\usepackage{nicefrac}       
\usepackage{microtype}      
\usepackage{xcolor}         
\usepackage[pdftex]{graphicx}
\usepackage{subcaption}
\usepackage{tabularx}
\usepackage{multirow}
\usepackage{natbib}
\setcitestyle{numbers}
\setcitestyle{square}

\title{Is Feedback All You Need? Leveraging Natural Language Feedback in Goal-Conditioned Reinforcement Learning}

\author{%
  Sabrina McCallum$^{1,2}$, Max Taylor-Davies$^{1,2}$, Stefano V. Albrecht$^{1}$, Alessandro Suglia$^{2}$\\
  \AND
  \textnormal{$^{1}$University of Edinburgh $^{2}$Heriot-Watt University}\\
}

\begin{document}

\maketitle
\begin{abstract}
Despite numerous successes, the field of reinforcement learning (RL) remains far from matching the impressive generalisation power of human behaviour learning. One possible way to help bridge this gap be to provide RL agents with richer, more human-like feedback expressed in natural language. To investigate this idea, we first extend BabyAI to automatically generate language feedback from the environment dynamics and goal condition success. Then, we modify the Decision Transformer architecture to take advantage of this additional signal. We find that training with language feedback either in place of or in addition to the return-to-go or goal descriptions improves agents’ generalisation performance, and that agents can benefit from feedback even when this is only available during training, but not at inference.\footnote{Code available at github.com/uoe-agents/feedback-dt}
\end{abstract}

\section{Introduction}
Despite significant advances over decades of research, modern AI systems still lag significantly behind the learning abilities of humans. During their development, human infants develop a wide range of adaptive behaviours through an open-ended learning process that is remarkable for both its sample efficiency and generalisation \cite{saffran1996statistical}. One likely contributing factor is that, while infant learners do engage in trial-and-error learning, they are also able to draw from a variety of feedback sources beyond their immediate environment. One such source is other humans, who can provide them with additional feedback signals in the form of natural language \cite{Hattie2007ThePO, Schulz2012TheOO}. Sometimes, this feedback is akin to a classic reward signal, such as a parent praising (or scolding) their child for doing something right (or wrong). But it can also be richer and more structured, conveying information tailored to the learner's current goal, such as explanations of specific observations or events. This allows the learner to update their prior knowledge and build a more stable and accurate model of the world \cite{miller2019explanation, lombrozo2006structure}. 

While they are able to achieve superhuman performance on tasks such as Go \cite{silver2016mastering, silver2017mastering}, Starcraft II \cite{vinyals2019grandmaster} and Dota   \cite{Berner2019Dota2W}, reinforcement learning (RL) agents, in contrast to humans, typically struggle with both sample efficiency and generalisation. This is seen especially in settings where the environment reward is sparse or under-specified \cite{ocana2023overview}. Additionally, there may be environment–task combinations for which no sufficiently expressive Markov reward function exists, or, for tasks specified in language, a mismatch in abstraction between task and reward could arise \cite{Abel2021OnTE}. Inspired by human learning, we therefore investigate whether RL agents can learn more generalisable policies in sparse-reward environments when introduced to richer and more human-like feedback signals expressed through natural language. We modify an existing offline RL algorithm, specifically the Decision Transformer \cite{Chen2021DecisionTR}, to condition on different types of language feedback provided by the environment. Our method includes a procedure to automatically generate this language feedback based on the agent's actions and current goal, and requires no human-in-the-loop involvement. We find evidence that conditioning on language feedback can boost generalisation performance relative to baselines conditioning only on goal instructions or returns, and can match or outperform these baselines when goal instructions or returns are not available, even when feedback is not provided at inference time.

\section{Preliminaries}

\paragraph{Offline reinforcement learning.} We apply our method in the context of offline RL and model the decision-making process as a Partially Observable Markov Decision Process (POMDP), which generalises the Markov Decision Process (MDP) to cases where the underlying states cannot be observed directly by the agent. The MDP tuple here consists of finite sets of unobservable states $S$ and actions $A$, a transition dynamics function $T(s, a, s') = P(s'|s, a)$, a reward function $r = R(s, a)$, a distribution $\mu$ of initial states $s_{0}$, a finite set of observations $\Omega$, an observation function $O(s', a, o') = P(o'|s', a)$, and a discount factor $\gamma \in [0, 1)$. The agent selects actions according to a policy $\pi(a|o)$, which, for the partially observable case, defines a distribution over actions conditioned on observations. The optimal policy to be learned is specified by a learning objective, most commonly the maximisation of the expected discounted return, or cumulative reward, $\mathbb{E}_{\pi}[\sum_{t=1}^{T}\gamma^{t}r_{t}]$. In offline RL problems, the agent only has access to a fixed dataset $D = \{(o^{(i)}_{t}, a^{(i)}_{t}, r^{(i)}_{t}, o^{(i)}_{t+1})\}$ of rollouts of trajectories generated using an unknown (sub-optimal) policy $\pi_{B}$, and is not permitted to explore the environment to collect additional data. We consider only the \emph{sparse-reward} setting, where the learning problem is made harder by the fact that the agent receives positive reward only upon successfully achieving its goal. We further assume that $D$ does \emph{not} contain the optimal behaviour, and that consequently, naive imitation would result in sub-optimal performance.

\paragraph{Decision Transformer.} Rather than
relying on past rewards, the Decision Transformer (DT) \cite{Chen2021DecisionTR} conditions action generation on future desired returns. This is modelled using returns-to-go, where $\hat{R_{t}}$ is the discounted return-to-go from timestep $t$ to the end of the episode $\hat{R_{t}} = \sum_{t'=t}^{T}\gamma^{t'-t}r_{t'}$. At test time, the behaviour can be specified by providing an appropriate desired return, such as 1 for success or 0 for failure, alongside the initial state. The DT approach leverages the GPT \cite{Radford2019LanguageMA} architecture, which achieves autoregressive generation by means of a causal self-attention mask, so that the prediction of the action at timestep $t$ depends only on tokens from timesteps up to $t$. For computational reasons, this is typically limited to a given context size which includes the last $K$ timesteps in the input to the model. Where $K = 1$, the resulting policy is considered Markovian. Trajectories are represented as sequences of return-to-go (RTG), action, and state $\tau = (\hat{R}_{1}, s_{1}, a_{1}, \hat{R}_{2}, s_{2}, a_{2}, \dots, \hat{R}_{T}, s_{T}, a_{T})$, as this format lends itself to training and generation in an autoregressive fashion. Instead of maximising the expected discounted return, the learning objective of the Decision Transformer is to minimise the next-action prediction loss given the history and current state, typically measured as either cross-entropy loss (for discrete actions) or mean-squared error (for continuous actions).

\section{Related work}
\paragraph{Reinforcement Learning from Human Feedback (RLHF).} A related area of work can be found in RLHF, which describes a methodology for dynamically adapting machine learning models, such as large language models (LLMs) \cite{Scheurer2023TrainingLM, Ouyang2022TrainingLM, Stiennon2020LearningTS, Ziegler2019FineTuningLM} and more recently, vision-and-language models (VLMs) \cite{yang2023octopus, Sun2023AligningLM}, to hard-to-specify goals which are typically linked to human preferences or behaviours.
Responses generated by the base model are typically rated or ranked by human evaluators, and this feedback is used to train a reward model via supervised learning. Finally, the learned reward model is used to iteratively refine the outputs of the base model to align more closely with users' expressed preferences \cite{Casper2023OpenPA}. Both RLHF and our work use auxiliary feedback information to aid learning where it is potentially not easily possible to capture the desired behaviour in a simple pre-specified reward signal. However, while in RLHF, learning from feedback typically occurs as a separate subsequent process, we consider it as part of the main training procedure. Likewise, we do not use the feedback samples to train any explicit reward model, but pass them directly to the core behaviour-learning model.
\clearpage
\paragraph{Feedback in LLM prompts.} 
Taking advantage of the in-context learning capabilities of pretrained LLMs, a number of recent studies on language tasks such as chain-of-thought reasoning \cite{An2023LearningFM, Peng2023CheckYF, Wang2023ShepherdAC, Madaan2022MemoryassistedPE} include language feedback provided by humans or other LLMs in model prompts. Concurrent work leverages feedback based on compiler errors in prompts for language-to-code generation \cite{chen2023teaching, Ni2023LEVERLT, Wang2023LeTILT, Yang2023InterCodeSA}. This is complemented by a growing body of work which prompts pretrained LLMs or VLMs to generate plans for a range of robotic manipulation and embodied AI tasks, and which incorporate feedback on the generated plan into the prompt. The feedback used is typically automatically generated and ranges from simple binary task completion feedback \cite{Huang2022InnerME} to more verbose feedback messages from the environment and execution errors \cite{Rana2023SayPlanGL, Skreta2023ErrorsAU, Wang2023VoyagerAO, yang2023octopus}. While these approaches inspired our notion of feedback and the mechanism used to generate it, we do not make use of pretrained models or in-context learning, and apply feedback specifically in the context of RL.

\begin{figure}[tb]
\centering
  \centering
  \includegraphics{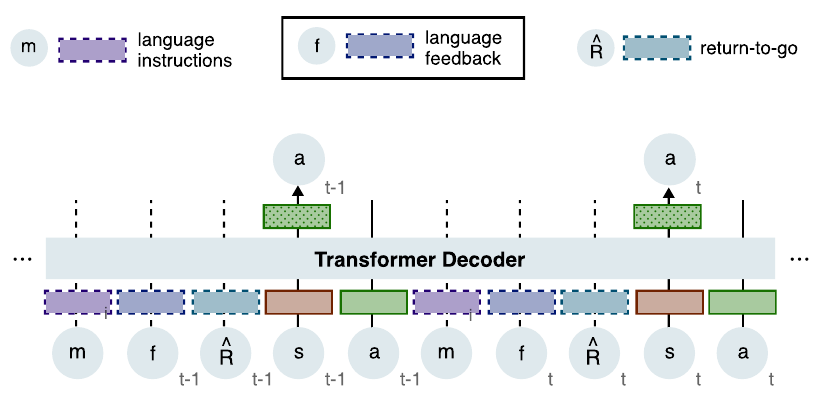}
\caption{We extend the capabilities of the Decision Transformer (return-conditioned) and Text Decision Transformer (instruction-conditioned), with the option to condition on language feedback.}
\label{fig:fdt}
\end{figure}

\section{Proposed method}
\label{section:method}
We propose the Feedback Decision Transformer (FDT), a method that leverages the Decision Transformer \cite{Chen2021DecisionTR} algorithm for language-conditioned offline RL and conditions action generation on sequences of actions, partial image observations and automatically generated language feedback, as illustrated in Figure \ref{fig:fdt}). The proposed architecture allows us to optionally combine language feedback with other signals proposed in previous studies, including returns-to-go \cite{Chen2021DecisionTR} and/or goal instructions \cite{putterman2022pretraining}, or rely soley on language feedback. The subsequent paragraphs describe our architecture, as well as the approach used to automatically generate language feedback from the learning environment, in more detail. 

\paragraph{Architecture.} We build upon the Decision Transformer \cite{Chen2021DecisionTR}, which casts RL as a sequence modelling problem, where behaviour is produced by generating action sequences in an auto-regressive manner and conditioned on the desired return. The Transformer \cite{Vaswani2017AttentionIA} has emerged as the architecture of choice for pre-training LLM's and VLM's, and has been shown to be competitive on offline RL and Imitation Learning (IL) benchmarks thanks to its flexibility with respect to input encoding, and its ability to condition on previous timesteps over long contexts through the self-attention mechanism. The Text Decision Transformer (TDT) \cite{putterman2022pretraining} adapts the original DT for goal-conditioned IL and conditions action generation on language goal instructions instead of RTG. Our architecture, the Feedback Decision Transformer (FDT), which is illustrated in Figure \ref{fig:fdt}, extends both methods by allowing action generation to be conditioned on RTG, goal instructions, language feedback, or any combination thereof. For the decoder, we adapt the implementation of the Decision Transformer based on GPT2 \cite{Radford2019LanguageMA} according to \cite{Chen2021DecisionTR}, and we train an encoder for the image observations concurrently with the decoder. Details are provided in Table \ref{tab:architecture} in Appendix \ref{appendix:model}. As in the original implementation, actions and RTG are encoded linearly, and we use absolute positional embeddings. We extend the original model with sentence-level embeddings for the mission and language feedback strings, for which we downsample pre-trained embeddings from a frozen SentenceBERT model \cite{Reimers2019SentenceBERTSE}. All embeddings are 128-dimensional, and language embeddings are provided one sentence per timestep. We minimise the mean cross-entropy loss across the discrete actions using the standard, unweighted reduction, as we find that computing a weighted mean of the losses based on the length of the episode achieves no significant improvements. We do not predict observations, returns-to-go, mission or language feedback. However, existing work on world models \cite{Ha2018WorldM}, such as Dreamer \cite{Hafner2019DreamTC}\cite{Hafner2020MasteringAW}\cite{Hafner2023MasteringDD}, TransDreamer \cite{Chen2022TransDreamerRL}, and most notably Dynalang \cite{Lin2023LearningTM}, which predicts not only future states and rewards, but also future language, shows this to be a promising avenue for future research.

\begin{figure}[tb]
\centering
  \centering
  \includegraphics{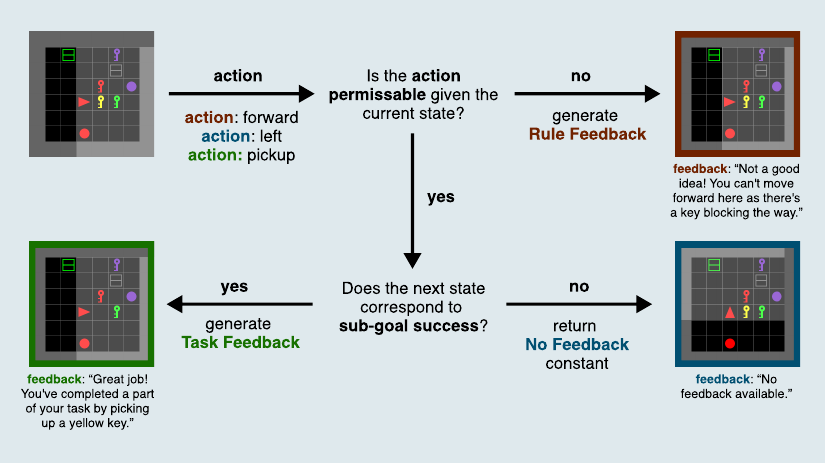}
\caption{Different actions can result in either Rule Feedback, no feedback, or Task Feedback.}
\label{fig:feedback}
\end{figure}

\paragraph{Augmenting environments with language feedback.} We propose a method to automatically generate low-level language feedback using predefined rules and templates, and define two feedback types, 'Rule Feedback' and 'Task Feedback', which capture different information about the consequences of the agent's actions, as is illustrated in Figure \ref{fig:feedback}. Note that according to our feedback generation procedure, the environment never returns both Rule and Task Feedback at the same timestep, and that not every action results in either of the feedback types to be generated. When neither Rule nor Task Feedback is triggered, we return the constant string "No feedback available.". For our Task Feedback, we decompose high-level goal instructions into granular sub-goals in order to generate feedback on the agent's progress towards the goal. The current sub-goal completed by the agent is then explicitly referred to in the Task Feedback string, along with a message that specifies whether an intermediate goal condition, or the final goal condition of the task has been met. Rule Feedback is provided when an action is executed that violates any of the physical constraints or predicates imposed by the environment. We conceptualise this as a type of corrective feedback extended with a detailed explanation for the failure. Note that while correction feedback is referenced in other work, such as HomeGrid \cite{Lin2023LearningTM}, our Rule Feedback does not rely on heuristics which are outside of the dynamics of the simulator, such as distance measures, and we avoid the use of instruction language in the feedback to clearly distinguish this from the goal instructions. In a similar vein to recent work in the space of LLMs for language-to-code generation \cite{yang2023octopus} and planning \cite{Rana2023SayPlanGL}, which leverages execution errors for actions as automatic feedback, we exploit the internal action validation logic for simulators that do not return execution errors or similar system messages as such. Due to the way we conceptualise Rule Feedback, it generally coincides with the previous observation being repeated, that is, the environment does not change as a result of the action. For templates and rules, please refer to Appendix \ref{appendix:datasets}.

\section{Experiments}
In our experiments, we test if an agent's ability to generalise to unseen environments and tasks when rewards are sparse can be improved when it has access to language feedback on its actions. Specifically, we investigate 1) whether different types of language feedback can complement traditional numerical feedback in the form of RTG, or goal instructions (mission strings) to improve performance, and 2) whether language feedback can replace RTG or goal instructions to achieve equivalent or improved performance. We consider whether the answer to these research questions differs for different types of tasks and environments, as well as generalisation scenarios, and when language feedback is provided only during training, but not at test time. Lastly, we test Task Feedback and Rule Feedback separately and in combination, and examine how this impacts 2).

\subsection{Experiment setup}

\paragraph{Environment and datasets}
To test our approach, we extend BabyAI \cite{ChevalierBoisvert2018BabyAIAP}, a suite of 2D gridworld environments which facilitates the training and benchmarking of agents on goal-oriented tasks specified in language. Levels in BabyAI increase in difficulty, and range from simple single rooms containing only the goal object, to complicated mazes with many objects. Harder levels involve multiple sub-goals, long horizons and complex compositional instructions. All levels in BabyAI pose challenges associated with sparse rewards, as only a terminal reward $r \in (0,1]$ is provided at the end of successful episodes, otherwise the reward is 0. In keeping with the original experiments in BabyAI in \cite{ChevalierBoisvert2018BabyAIAP}, we discount the returns using $\gamma = 0.99$, which is aligned with the sparse-reward setting. We select a subset of eight levels including four single rooms and four mazes, which are listed in Table \ref{tab:levels}, and use a random policy to generate offline training datasets for each level. For the single-room levels, we generate 10 different trajectories for 128 environment instances each, and 10 times as many for the more complex maze levels. Trajectories are composed of sequences of mission string, partial image observation, discrete action, scalar reward and feedback string for each timestep. Note that we include trajectories independently of whether the episode was successful or not, and that harder levels contain potentially very few successful episodes. Since our training datasets do not contain optimal trajectories, our setting differs from both those of the original DT and the TDT.

\newcolumntype{a}{X}
\newcolumntype{j}{>{\hsize=.5\hsize}X}
\newcolumntype{k}{>{\hsize=.4\hsize}X}
\newcolumntype{d}{>{\hsize=.3\hsize}X}
\newcolumntype{f}{>{\hsize=.25\hsize}X}
\newcolumntype{e}{>{\hsize=.2\hsize}X}
\newcolumntype{z}{>{\hsize=.15\hsize}X}

\begin{table}[t!]
    \centering
    \caption{BabyAI levels used in our experiments. The top four levels are single rooms, the bottom four mazes. GC's = goal conditions. Only GoToObj has no distractor objects. *Cannot be "door".}
    \begin{tabularx}{\textwidth}{daeee}
    \toprule
       Level & Mission space & GC's & Steps$_{max}$ & Episodes\\
       \midrule
       GoToObj & \verb|go to {the/a} {col} {type}| & 1 & 64 & 1,280\\
       GoToLocal & \verb|go to {the/a} {col} {type}| & 1 & 64 & 1,280\\
       PickupLoc & \verb|pick up {the/a} {col} {type}* {loc}| & 2 & 64 & 1,280\\
       PutNextLocal & \verb|put {the/a} {col1} {type1}| \verb|next to {the/a} {col2} {type2}| & 4 & 128 & 1,280\\
       \midrule
       Pickup & \verb|pick up {the/a} {col} {type}*| & 2 & 576 & 12,800\\
       PutNext & \verb|put {the/a} {col1} {type1}| \verb|next to {the/a} {col2} {type2}| & 4 & 1,152 & 12,800\\
       Synth & \verb|go to {the/a} {col} {type}|, \verb|pick up {the/a} {col} {type}|, \verb|open {the/a} {col} door|, \verb|put {the/a} {col1} {type1}| \verb|next to {the/a} {col2} {type2}| & 1/2/4 & 1,152 & 12,800\\
       SynthLoc & Same as Synth, but with \verb|loc| language & 1/2/4 & 1,152 & 12,800\\
       \bottomrule
    \end{tabularx}
    \label{tab:levels}
\end{table}

\paragraph{Performance measure.} By default, BabyAI environments include high-level language goal instructions, or 'missions', which consist of one or multiple action instructions (either \verb|go to|, \verb|open|, \verb|pick up|, or \verb|put next|) paired with goal objects, which are specified using color, type and location descriptors. Apart from the simple \verb|goto|, these actions involve multiple steps; we decompose them into their component steps, which serve as sub-goals (or goal conditions) for the purpose of generating Task Feedback and calculating the goal-condition success rate \cite{Shridhar2019ALFREDAB}, which we use to evaluate and compare model performance. Goal-condition success rate is commonly used in instruction following tasks \cite{Shridhar2019ALFREDAB, Akula2022ALFREDLIT, Padmakumar2021TEAChTE, Shridhar2020ALFWorldAT} and allows for a more granular, non-binary perspective on task success. We calculate the final score as the average goal-condition success of each test episode. To illustrate our conceptualisation of goal conditions, the mission "put a yellow ball next to the green box" would be decomposed into four sub-goals: \verb|go to a yellow ball|, \verb|pick up a yellow ball|, \verb|go next to the green box|, and \verb|put a yellow ball next to the green box|. Note that we do not use these sub-goals directly as a learning signal.

\paragraph{Zero-shot compositional generalisation.} We devise an evaluation protocol for zero-shot generalisation inspired by previous work on compositional generalisation for grounded language learning, including benchmarks such as CompGuessWhat?! \cite{Suglia2020CompGuessWhatAM}, as well as gSCAN \cite{Ruis2020ABF} and its descendants \cite{Cao2020ZeroShotCP, HeinzeDeml2020ThinkBY, Wu2021ReaSCANCR, Sikarwar2022WhenCT, Aghzal2023CanLL}, which, while predominantly explored in the context of supervised learning, is relevant for work in RL \cite{Kirk2021ASO}. Specifically, we design three combinatorial interpolation \cite{Kirk2021ASO} scenarios, and two extrapolation scenarios. The combinatorial interpolation scenarios test the agent's ability to generalise to novel combinations of context dimension values that have been seen individually during training, but not together \cite{Kirk2021ASO, hupkes2020compositionality}. For instance, an agent that was trained to solve missions where the possible set of goal objects would have included yellow keys and green boxes, but never yellow boxes, would be required to systematically recombine the values "yellow" for the goal object colour dimension and "box" for the goal object type dimension at test time. For tasks where the instructions involve putting one goal object next to another (fixed) goal object, we hold out type-value combinations for either the first or the second object. In addition, we test an unseen combination of X and Y coordinates for the room quadrant the agent is spawned in. Extrapolation scenarios test the agent's ability generalise to values outside of the range of values that were seen during training for a given context dimension. We apply this to a categorical dimension (the relative location of the goal object to the agent's starting position) and a discrete numerical dimension (the size of rooms). We evaluate generalisation on environments seeded with 128 held-out seeds each for in-distribution (IID) and out-of-distribution (OOD) contexts, whereby the IID seeds are drawn from the same distribution as, but do not include, the training seeds, and both training seeds and IID test seeds seed missions that do not contain the held-out values for our interpolation and extrapolation scenarios described above. The OOD environments are OOD with respect to exactly one of the scenarios. We report OOD performance averaged across the different scenarios, as well as broken down by scenario, in Appendix \ref{appendix:results}.

\begin{table}[t!b]
    \centering
    \caption{Attributes and held-out values used to evaluate OOD generalisation in our experiments. *'Fixed goal object' is only applicable to levels where the mission space includes PutNext instructions. **'Relative goal location' is only applicable to levels with location language (\emph{Loc} levels).}
    \begin{tabular}{lll}
    \toprule
        & Scenario  &  Held out\\
       \midrule
       \multirow{3}{*}{Interpolation} & Goal object colour and type & yellow box \\
       & Fixed goal object* colour and type & blue ball \\
       & Agent starting location in room & top left\\
       \midrule
       \multirow{2}{*}{Extrapolation} & Relative goal location** & on your right \\
       & Room size & < 8 x 8 tiles \\

       \bottomrule
    \end{tabular}
    \label{tab:ood}
\end{table}

\paragraph{Model variants and ablations.} We compare the relative performance of different variants of our feedback-conditioned model against baselines which condition only on RTG or only on mission, and investigate a) whether conditioning on language feedback in addition to RTG or mission boosts the generalisation performance of the baselines, b) whether we can exceed, or at least match, the generalisation performance of the baselines by relying solely on language feedback, and whether this differs across generalisation scenarios and levels. Additionally, we investigate c) whether providing feedback not only during training but additionally at inference can improve generalisation performance. This means that, while for variants using the mission string, we provide the mission string for both training and inference, for those variants using feedback, we ablate whether the model has access to the actual feedback at inference time or whether it is given a constant, randomly sampled placeholder embedding. For models using the RTG, we use a target RTG of 1 during inference, which is the maximum achievable in BabyAI. We use a context length of up to 64 timesteps. Further details on the training procedure are provided in Appendix \ref{appendix:model}.

\begin{figure}[t!]
\centering
\begin{subfigure}{.5\textwidth}
  \centering
  \includegraphics{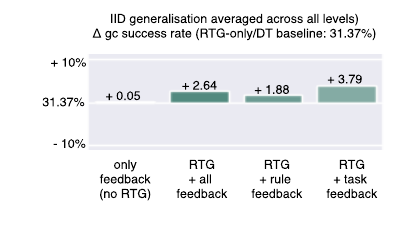}
  \label{fig:results-rtg-iid}
\end{subfigure}%
\begin{subfigure}{.5\textwidth}
  \centering
  \includegraphics{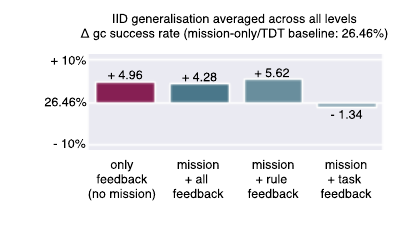}
  \label{fig:results-mission-iid}
\end{subfigure}
\begin{subfigure}{.5\textwidth}
  \centering
\includegraphics{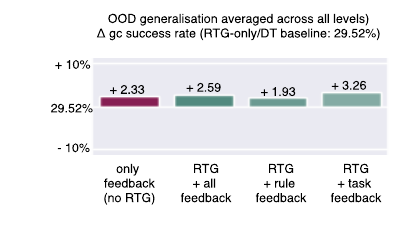}
  \label{fig:results-rtg-ood}
\end{subfigure}%
\begin{subfigure}{.5\textwidth}
  \centering
  \includegraphics{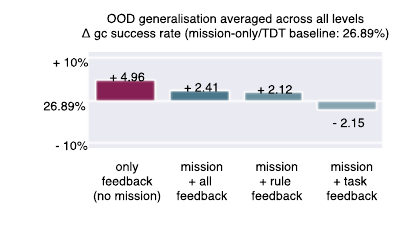}
  \label{fig:results-mission-ood}
\end{subfigure}
\caption{Difference in performance between the return-only (DT) and mission-only (TDT) baselines and our proposed variants that condition with feedback. For all results, including a breakdown by level and OOD type, please refer to Appendix \ref{appendix:results}.}
\label{fig:fig3}
\end{figure}

\subsection{Experimental results}
Our empirical results indicate that for tasks in the sparse-reward, partially observable setting of BabyAI, conditioning action generation on language feedback can facilitate the learning of policies that generalise more successfully than when action generation is conditioned solely on RTG or one-off language instructions (i.e., the mission string). While in some cases, language feedback alone is an insufficient learning signal, we find that it can boost performance when provided in addition to RTG or language goal instructions, the other signals considered in our setup. Specifically, we observe that language feedback tends to be more useful for certain task and environment types, and supports certain generalisation scenarios more so than others. Additionally, our results suggest that the two types of feedback - Task and Rule Feedback - do not necessarily complement each other and are not always equally suited to be used in combination with the RTG and mission. The following paragraphs provide more details on these observations.

\textbf{Combining RTG with Task Feedback boosts performance, especially on single-room levels.}
Since Rule Feedback seems to negatively impact performance on most of the maze levels when combined with the RTG, we hypothesise that while this dense feedback discourages unnecessary behaviour that has no effect on the environment, thereby reducing the risk of the environment timing out in easier levels with lower max steps, it is less helpful in harder maze levels, where the agent is allowed to try to complete the task for longer, and potentially distracts from the goal-relevant behaviour encouraged by the RTG. 

\textbf{Combining mission with Rule Feedback boosts performance, especially on levels with longer horizons.} Conversely, we observe that combining the mission with only the Task Feedback can be detrimental for performance. Unlike Rule Feedback, which is explicitly linked to the effect of the previous action, Task Feedback encapsulates the cumulative effect of the full sequence of actions since the previous goal condition, and we hypothesise that when presented with the mission and the Task Feedback together, the model has to learn to simultaneously leverage two long-horizon signals at the same time.

\textbf{Replacing RTG with feedback improves OOD performance, especially when extrapolating to new goal locations.}
We record performance improvements for all language-related OOD scenarios, and find that the improvements are more significant the lower the performance of the RTG-only baseline is, as well as when the relative goal location is OOD. These findings could be an indication that language feedback captures transferable, higher-level information where behaviour learned on the basis of a numerical reward is perhaps too specific to the training configurations and tasks.

\textbf{Replacing mission with feedback improves OOD performance, especially on levels with two or more sub goals.} We observe the most significant improvements when the performance of the baseline is poor and the model seemingly fails to learn a generalisable mapping to task-relevant behaviour from the mission alone, as well as for test environments where the model has to generalise to unseen relative goal locations or fixed goal objects, both of which are specified in the final part of the mission. 

\textbf{Providing feedback at inference is only useful when performance is otherwise poor.} Apart from the feedback-only variant, the performance change with feedback at inference appears to be inversely proportional to the performance achieved without feedback at inference. While this behaviour merits further investigation, we hypothesise that this is possibly due to a shift in the distribution of feedback encountered at inference compared with the distribution of feedback in the training episodes, and that this negatively affects the agent's performance.

\section{Limitations and future work}
While we limit ourselves to demonstrating the potential of language feedback for goal-conditioned RL on a single algorithm and learning environment, we are optimistic that the underlying idea can be transferred successfully to other RL algorithms and environments beyond 2D gridworlds---we aim to explore this in future work. Replacing templates with more diverse language, e.g. generated with LLM prompting, similar to recent work in Reinforcement Learning from AI Feedback (RLAIF) \cite{Lee2023RLAIFSR}, could provide the additional flexibility and scalability required for transfer to more diverse task, action and observation spaces, without having to rely on humans in the loop. 

As our feedback is generated automatically and does not rely on human annotators, it should translate directly to the online setting in the scope of simulated learning environments. It remains to be seen if pre-training agents in this or a similar fashion would be sufficient for successful sim-to-real transfer; however, we believe that providing free-form language feedback to robots deployed in the real world would be intuitive for human collaborators and operators. Exploring the implications of this approach to Human-Robot-Interaction is an interesting and promising avenue for future research.

Our approach is multi-modal insofar as we utilise both image and language inputs, but the representations for the different modalities are learned in isolation from one another. Future work could see this extended to multi-modal representations that are both grounded and situated.

\section{Conclusion}
We investigate the potential of using automatically generated language feedback to train agents in sparse-reward environments with language-specified goals. We find evidence that conditioning on such feedback in addition to goal instructions or desired return can yield significant improvements in generalisation to unseen environments, including environments that correspond to one of multiple OOD scenarios and that require agents to interpolate or extrapolate to new contexts. Within the BabyAI environment suite, feedback seems to provide a useful complementary signal to desired return in easy levels, and to goal instructions in harder levels. Additionally, we establish that language feedback can potentially serve as an alternative condition when goal instructions or desired return are not available. Lastly, we observe that while in most cases, providing feedback during training is sufficient, there are instances where feedback is only effective when it is also provided at test time.

\clearpage
\bibliographystyle{IEEEtran}

\bibliography{references} 

\clearpage

\appendix

\section{Appendix}

\subsection{Datasets}
\label{appendix:datasets}
\paragraph{BabyAI levels} We select only a subset of the original levels in BabyAI. For an exhaustive overview of the remaining levels, the reader may refer to the original paper \cite{ChevalierBoisvert2018BabyAIAP}. We use configurations of these levels that are registered as Gymnasium \cite{towers_gymnasium_2023}. Some levels have multiple registered configurations, which differ in terms of room size, the number of distractor objects, closed doors, etc. In the case of levels with multiple configurations, we use all available configurations, although we consider some configurations to be entirely OOD, usually with respect to the room size. Note that the object \verb|col| dimension can take one of six values ("blue", "green", "grey", "red", "purple", "yellow"), and the \verb|type| dimension one of four ("ball", "box", "key", "door"), and \verb|loc| can be either "on your left", "on your right", "above" or "below". Note that levels which include the \verb|pick up| action cannot be instantiated with \verb|type| "door", and that location language (\verb|loc|) is relative to the agents starting position and is not updated to reflect the agent's movement. 

\paragraph{Feedback generation} Note that our feedback generation is entirely deterministic, and we fully control language through the use of hand-crafted templates, rather than by prompting an LLM. Feedback is generated when the corresponding rule is triggered. For Rule Feedback, this refers to certain combinations of the action and a pre-condition. In the case of Task Feedback, the rule corresponding to sub-goal success is determined by the instruction type. Both sets of rules and the corresponding templates are provided in Tables \ref{tab:task-feedback-templates} and \ref{tab:rule-feedback-templates}.

\begin{table}[h!]
    \centering
    \caption{Task Feedback templates and their corresponding rules, which check whether the effect of an action corresponds to the current goal condition. Note that unlike the other instruction types, GoNextTo is not a default BabyAI instruction type and used exclusively as a sub-goal for PutNext instructions.}
    \begin{tabularx}{\textwidth}{feja}
    \toprule
       Instruction type & Action & Condition & Feedback template\\
       \midrule
       \verb|GoTo| & \verb|forward| \verb|left| \verb|right| & Goal(FrontCell) & "Fantastic! You've completed \{a part of \}your task by going to \{goal object description\}."\\
       \midrule
       \verb|GoNextTo| & \verb|forward| \verb|left| \verb|right| & NextTo(FrontCell, Goal) & "That's right! You've completed \{a part of \}your task by going next to {goal object description}."\\
       \midrule
       \verb|Open| & \verb|toggle| & Goal(FrontCell) $\land$ OpenDoor(FrontCell) & "That's correct! You've completed \{a part of \}your task by opening \{goal door description\}." \\
       \midrule
       \verb|Pickup| & \verb|pickup| & Carrying(Object) $\land$ Goal(Object) & "Great job! You've completed \{a part of \}your task by picking up \{goal object description\}."\\
       \midrule
       \verb|PutNext| & \verb|drop| & NextTo(FrontCell, FixedGoal) $\land$ MoveGoal(FrontCell) & "That's right! You've completed \{a part of \}your task by going next to {goal object description}."\\
       \bottomrule
    \end{tabularx}
    \label{tab:task-feedback-templates}
\end{table}

\begin{table}[htb]
    \centering
    \caption{Rule Feedback templates and their corresponding rules, which check whether an action has violated any of the pre-conditions for the action to have an effect on the environment.}
    \begin{tabularx}{\textwidth}{eka}
    \toprule
       Action & Condition & Feedback template\\
       \midrule
       \verb|forward| & Wall(FrontCell) & "Not a good idea! You can't move forward while you're facing the wall."\\
       \verb|forward| & Object(FrontCell) $\land$ $\lnot$Door(FrontCell) & "Not a good idea! You can't move forward here as there's a \{object\} blocking the way."\\
       \verb|forward| & Door(FrontCell) $\land$ Closed(FrontCell) & "Not a good idea! You can't move forward here as the door in front of you is closed." \\
       \verb|forward| & Door(FrontCell) $\land$ Locked(FrontCell) & "Not a good idea! You can't move forward here as the door in front of you is locked."\\
       \midrule
       \verb|pickup| & Wall(FrontCell) & "Not a good idea! You can't pick up the wall."\\
       \verb|pickup| & Empty(FrontCell) & "Not a good idea! There's nothing in front of you, and you can't pick up empty space."\\
       \verb|pickup| & Door(FrontCell) & "Not a good idea! You can't pick up doors."\\
       \verb|pickup| & Object(Carrying) & "Not a good idea! You can't pick up another object while you're already carrying one."\\
       \midrule
       \verb|drop| & Wall(FrontCell) & "Don't do that! You can't drop an object while you're facing the wall."\\
       \verb|drop| & Object(FrontCell) $\land$ $\lnot$Door(FrontCell) & "Don't do that! You can't drop an object on top of another object, and there's already a \{object type\} in front of you." \\
       \verb|drop| & Door(FrontCell) & "Don't do that! You can't drop an object while you're facing a door."\\
       \verb|drop| & Empty(Carrying) & "Don't do that! You're not carrying any object so dropping has no effect." \\
       \midrule
       \verb|toggle| & Wall(FrontCell) & "That won't work here. You can't open the wall."\\
       \verb|toggle| & Object(FrontCell) $\land$ $\lnot$Box(Object) & "That won't work here. You can't open \{object type\}s."\\
       \verb|toggle| & Empty(FrontCell) & "That won't work. There's nothing in front of you, and you can't open empty space."\\
       \verb|toggle| & Door(FrontCell) $\land$ Locked(FrontCell) $\land$ \linebreak $\lnot$Key(Carrying) & "That won't work here. You can't open a locked door without a key of the same color as the door, and you're not carrying any key."\\
       \verb|toggle| & Door(FrontCell) $\land$ Locked(FrontCell) $\land$ \linebreak Key(Carrying) $\land$ $\lnot$SameCol(Carrying, FrontCell)   & "That won't work here. You can't open a locked door without a key of the same color as the door. You're carrying a \{key color\} key, but the door in front of you is \{door color\}."\\
       \bottomrule
    \end{tabularx}
    \label{tab:rule-feedback-templates}
\end{table}

\clearpage
        
\subsection{Model Architecture and training}
\label{appendix:model}
We keep most of the default values for training parameters listed in Table \ref{tab:training} as used by \cite{Chen2021DecisionTR} for their model corresponding to the GPT2-based architecture. We train for up to 10 epochs with early stopping based on goal-condition success on the training seeds, and using batches of 64 samples consisting of sub-episodes with context length 64. In the case of single-room levels with simple actions (\verb|goto| and \verb|pickup|), this corresponds to the full episode for unsuccessful episodes; successful episodes in these levels are shorter than our (maximum) context length. Note that we use random starting points within episodes from which to sample sub-episodes and that consequently, even sub-episodes sampled from episodes that are longer than the context length can contain fewer than 64 steps. The value used for early stopping patience is based on level complexity, whereby we double the patience value after first two levels and again after the first four levels.

\begin{table}[h]
    \caption{Architecture hyperparameters}
    \begin{subtable}{.5\linewidth}
      \centering
        \caption{Decoder \cite{Chen2021DecisionTR}}
        \label{tab:decoder}
        \begin{tabular}{ll}
        \toprule
            Parameter & Value\\
            \midrule
            $N$ layers & 3\\
            $N$ attention heads & 1\\
            Hidden dimension & 128\\
            Dropout & 0.1\\
            Non-linearity & ReLU\\
        \bottomrule
        \end{tabular}
    \end{subtable}%
    \begin{subtable}{.5\linewidth}
      \centering
        \caption{Image encoder \cite{ChevalierBoisvert2018BabyAIAP}}
        \label{tab:state-encoder}
        \begin{tabular}{ll}
        \toprule
            Parameter & Value\\
            \midrule
            $N$ convolutions & 3 \\
            Channels & 16, 32, 64 \\
            Filter sizes & 2x2, 2x2, 2x2\\
            Strides & 1, 1, 1\\
            Non-linearity & ReLU\\
        \bottomrule
        \end{tabular}
    \end{subtable} 
    \label{tab:architecture}
\end{table}

\begin{table}[h]
    \caption{Training hyperparameters}
    \begin{subtable}{.5\linewidth}
      \centering
        \caption{Optimisation}
        \begin{tabular}{ll}
        \toprule
            Parameter & Value\\
            \midrule
            Optimiser & AdamW\\
            $\beta_1$, $\beta_2$ & 0.9, 0.99\\
            Weight decay & $1\times10^-4$\\
            Learning rate & $5\times10^-4$\\
            Scheduler & Linear\\
            Warm-up ratio & 0.1\\
        \bottomrule
        \end{tabular}
    \end{subtable}%
    \begin{subtable}{.5\linewidth}
      \centering
        \caption{Other}
        \begin{tabular}{ll}
        \toprule
            Parameter & Value\\
            \midrule
            Max gradient norm & 0.25\\
            Max epochs & 10\\
            Early stopping &\\
            - patience (val steps) & 8/16/32\\
            - threshold & 0.01\\
            Batch size & 64 \\
        \bottomrule
        \end{tabular}
    \end{subtable} 
    \label{tab:training}
\end{table}
\vfill
\clearpage

\subsection{Further results}
\label{appendix:results}

\subsubsection{Performance by environment type}

\begin{figure}[htb]
\centering
\begin{subfigure}{.5\textwidth}
  \centering
  \includegraphics{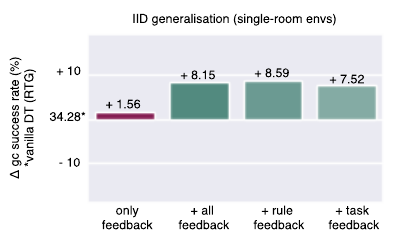}
\end{subfigure}%
\begin{subfigure}{.5\textwidth}
  \centering
  \includegraphics{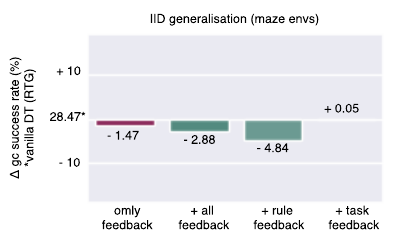}
\end{subfigure}
\begin{subfigure}{.5\textwidth}
  \centering
  \includegraphics{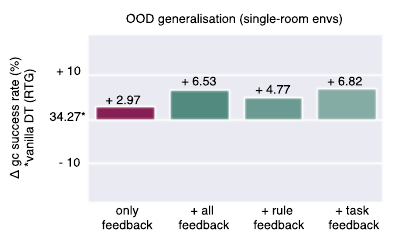}
\end{subfigure}%
\begin{subfigure}{.5\textwidth}
  \centering
  \includegraphics{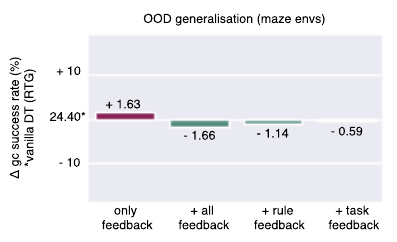}
\end{subfigure}
\caption{IID and OOD generalisation performance by environment type for the proposed variants conditioning on RTG and/or feedback compared against the RTG-only (vanilla DT) baseline.}
\end{figure}

\begin{figure}[h!]
\centering
\begin{subfigure}{.5\textwidth}
  \centering
  \includegraphics{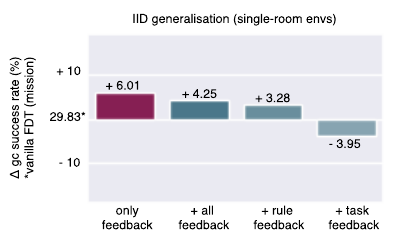}
\end{subfigure}%
\begin{subfigure}{.5\textwidth}
  \centering
  \includegraphics{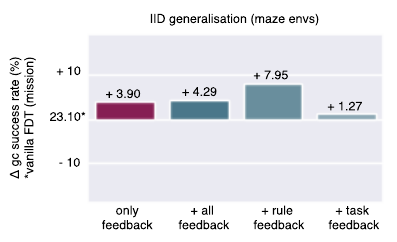}
\end{subfigure}
\begin{subfigure}{.5\textwidth}
  \centering
  \includegraphics{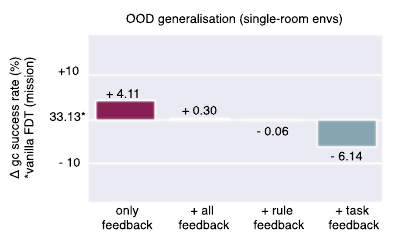}
\end{subfigure}%
\begin{subfigure}{.5\textwidth}
  \centering
  \includegraphics{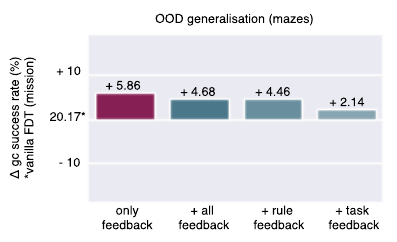}
\end{subfigure}
\caption{IID and OOD generalisation performance by environment type for the proposed variants conditioning on mission and/or feedback compared against the mission-only (vanilla TDT) baseline.}
\end{figure}

\clearpage

\subsubsection{Performance by level}

\begin{figure}[htb]
\centering
\begin{subfigure}{.5\textwidth}
  \centering
  \includegraphics{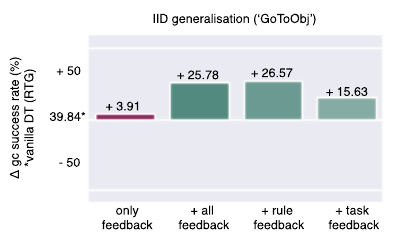}
\end{subfigure}%
\begin{subfigure}{.5\textwidth}
  \centering
  \includegraphics{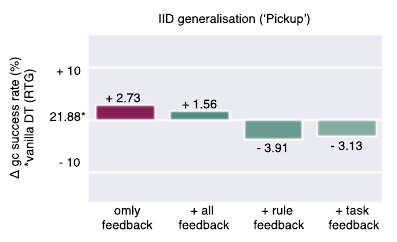}
\end{subfigure}
\begin{subfigure}{.5\textwidth}
  \centering
  \includegraphics{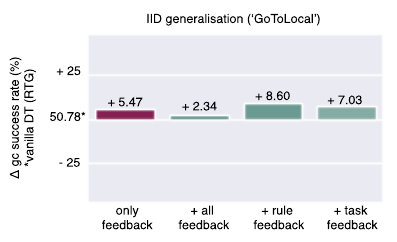}
\end{subfigure}%
\begin{subfigure}{.5\textwidth}
  \centering
  \includegraphics{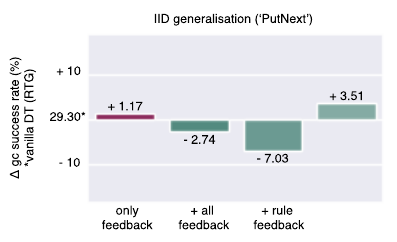}
\end{subfigure}
\begin{subfigure}{.5\textwidth}
  \centering
  \includegraphics{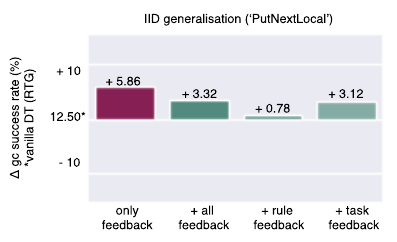}
\end{subfigure}%
\begin{subfigure}{.5\textwidth}
  \centering
  \includegraphics{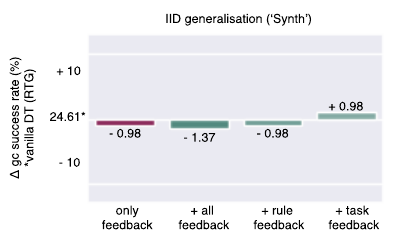}
\end{subfigure}
\begin{subfigure}{.5\textwidth}
  \centering
  \includegraphics{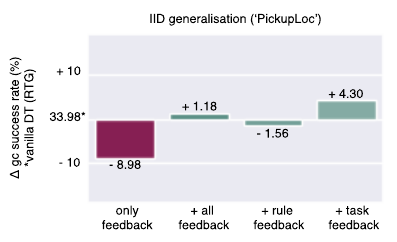}
\end{subfigure}%
\begin{subfigure}{.5\textwidth}
  \centering
  \includegraphics{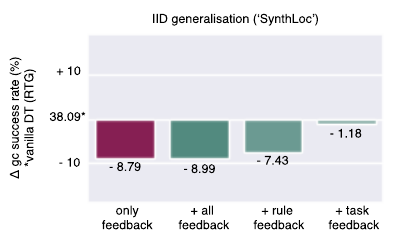}
\end{subfigure}
\caption{IID generalisation performance by level for the proposed variants conditioning on return and/or feedback compared against the return-only (vanilla DT) baseline.}
\end{figure}

\begin{figure}[htb]
\centering
\begin{subfigure}{.5\textwidth}
  \centering
  \includegraphics{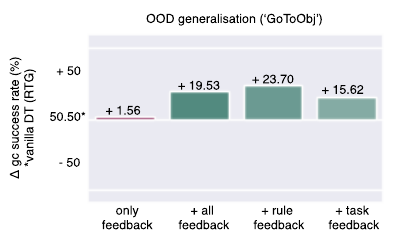}
\end{subfigure}%
\begin{subfigure}{.5\textwidth}
  \centering
  \includegraphics{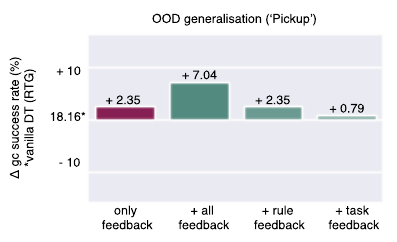}
\end{subfigure}
\begin{subfigure}{.5\textwidth}
  \centering
  \includegraphics{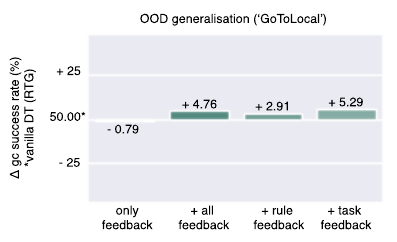}
\end{subfigure}%
\begin{subfigure}{.5\textwidth}
  \centering
  \includegraphics{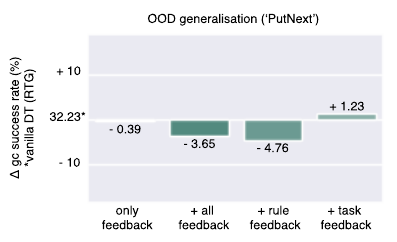}
\end{subfigure}
\begin{subfigure}{.5\textwidth}
  \centering
  \includegraphics{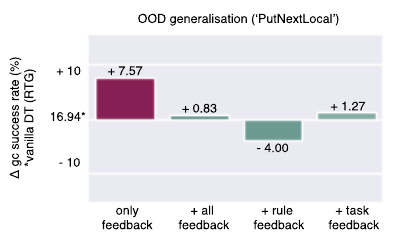}
\end{subfigure}%
\begin{subfigure}{.5\textwidth}
  \centering
  \includegraphics{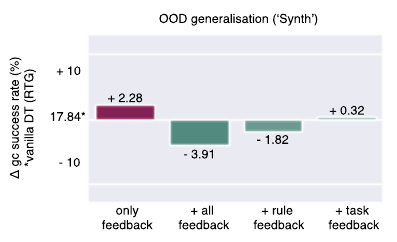}
\end{subfigure}
\begin{subfigure}{.5\textwidth}
  \centering
  \includegraphics{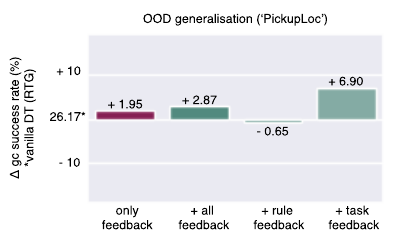}
\end{subfigure}%
\begin{subfigure}{.5\textwidth}
  \centering
  \includegraphics{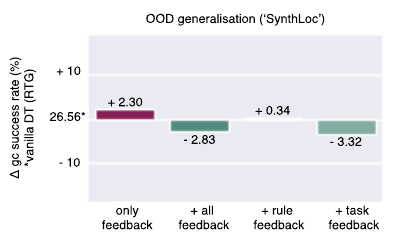}
\end{subfigure}
\caption{OOD generalisation performance by level for the proposed variants conditioning on return and/or feedback compared against the return-only (vanilla DT) baseline.}
\end{figure}

\begin{figure}[htb]
\centering
\begin{subfigure}{.5\textwidth}
  \centering
  \includegraphics{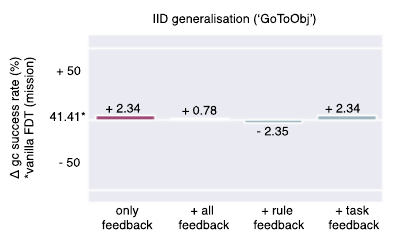}
\end{subfigure}%
\begin{subfigure}{.5\textwidth}
  \centering
  \includegraphics{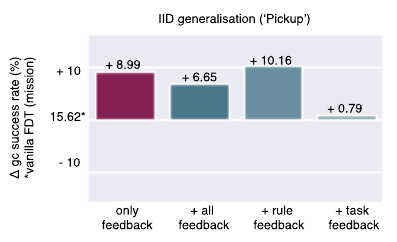}
\end{subfigure}
\begin{subfigure}{.5\textwidth}
  \centering
  \includegraphics{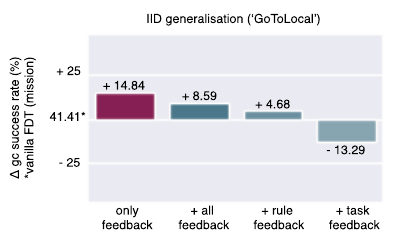}
\end{subfigure}%
\begin{subfigure}{.5\textwidth}
  \centering
  \includegraphics{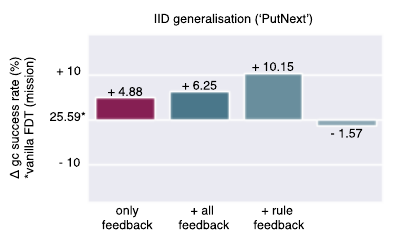}
\end{subfigure}
\begin{subfigure}{.5\textwidth}
  \centering
  \includegraphics{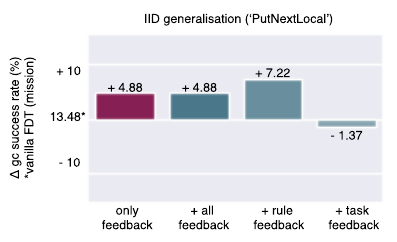}
\end{subfigure}%
\begin{subfigure}{.5\textwidth}
  \centering
  \includegraphics{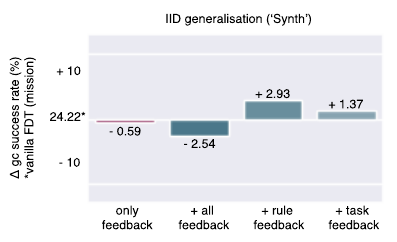}
\end{subfigure}
\begin{subfigure}{.5\textwidth}
  \centering
  \includegraphics{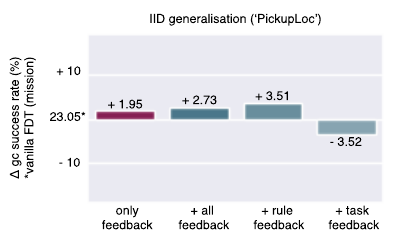}
\end{subfigure}%
\begin{subfigure}{.5\textwidth}
  \centering
  \includegraphics{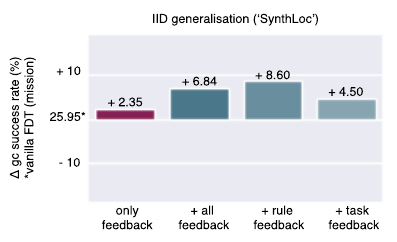}
\end{subfigure}
\caption{IID generalisation performance by level for the proposed variants conditioning on mission and/or feedback compared against the mission-only (vanilla TDT) baseline.}
\end{figure}

\begin{figure}[htb]
\centering
\begin{subfigure}{.5\textwidth}
  \centering
  \includegraphics{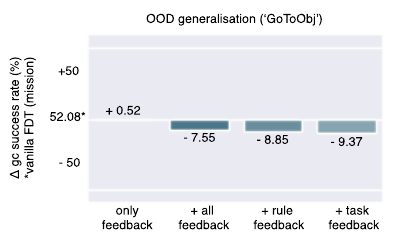}
\end{subfigure}%
\begin{subfigure}{.5\textwidth}
  \centering
  \includegraphics{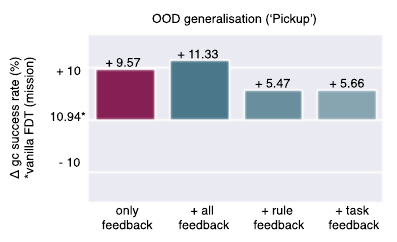}
\end{subfigure}
\begin{subfigure}{.5\textwidth}
  \centering
  \includegraphics{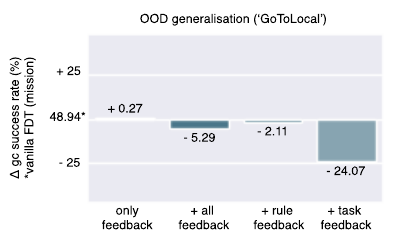}
\end{subfigure}%
\begin{subfigure}{.5\textwidth}
  \centering
  \includegraphics{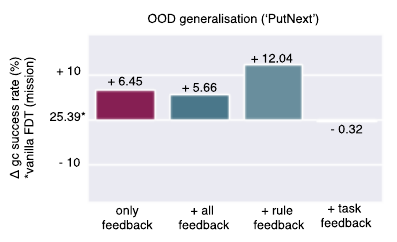}
\end{subfigure}
\begin{subfigure}{.5\textwidth}
  \centering
  \includegraphics{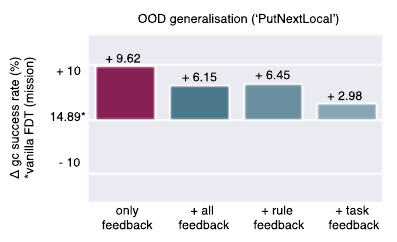}
\end{subfigure}%
\begin{subfigure}{.5\textwidth}
  \centering
  \includegraphics{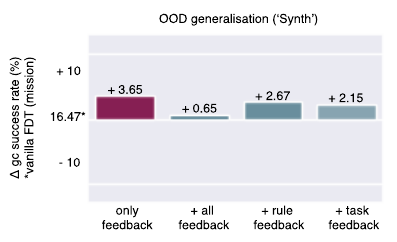}
\end{subfigure}
\begin{subfigure}{.5\textwidth}
  \centering
  \includegraphics{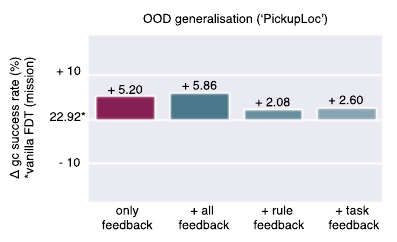}
\end{subfigure}%
\begin{subfigure}{.5\textwidth}
  \centering
  \includegraphics{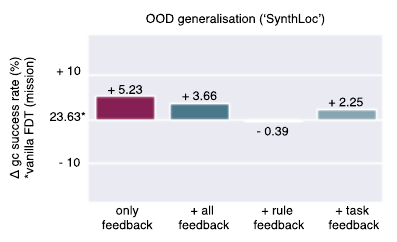}
\end{subfigure}
\caption{OOD generalisation performance by level for the proposed variants conditioning on mission and/or feedback compared against the mission-only (vanilla TDT) baseline.}
\end{figure}

\clearpage
\subsubsection{OOD performance by OOD type}

\begin{figure}[htb]
\centering
\begin{subfigure}{.5\textwidth}
  \centering
  \includegraphics{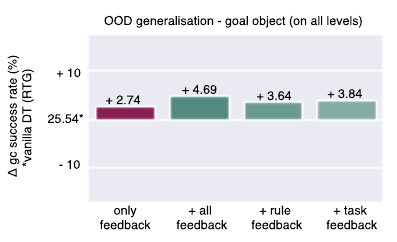}
\end{subfigure}%
\begin{subfigure}{.5\textwidth}
  \centering
  \includegraphics{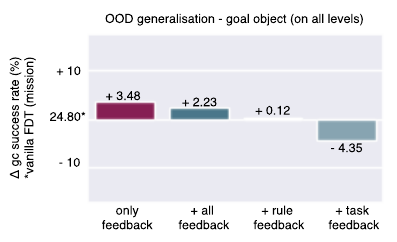}
\end{subfigure}
\begin{subfigure}{.5\textwidth}
  \centering
  \includegraphics{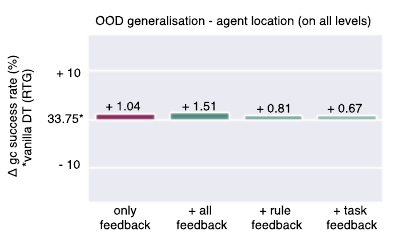}
\end{subfigure}%
\begin{subfigure}{.5\textwidth}
  \centering
  \includegraphics{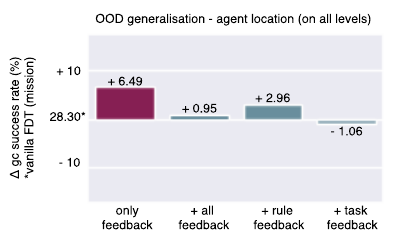}
\end{subfigure}
\begin{subfigure}{.5\textwidth}
  \centering
  \includegraphics{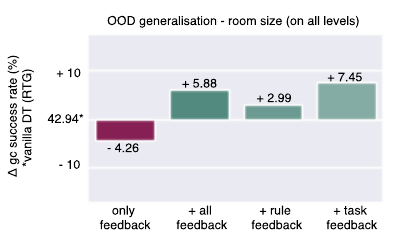}
\end{subfigure}%
\begin{subfigure}{.5\textwidth}
  \centering
  \includegraphics{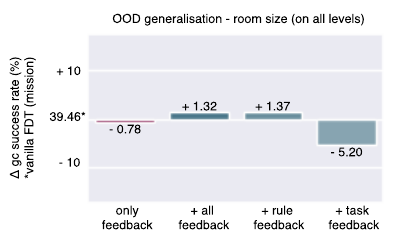}
\end{subfigure}
\begin{subfigure}{.5\textwidth}
  \centering
  \includegraphics{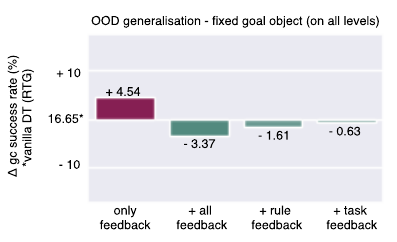}
\end{subfigure}%
\begin{subfigure}{.5\textwidth}
  \centering
  \includegraphics{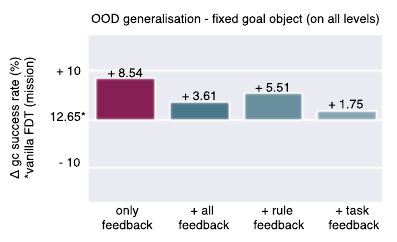}
\end{subfigure}
\caption{OOD generalisation performance across all eight levels for the proposed variants conditioning on return and/or feedback and mission and/or feedback compared against the return-only (vanilla DT) and mission-only (vanilla TDT) baselines, respectively, w.r.t. agent starting location, goal object color and type, room size and fixed goal object color and type.}
\end{figure}

\clearpage

\begin{figure}[htb]
\centering
\begin{subfigure}{.5\textwidth}
  \centering
\includegraphics{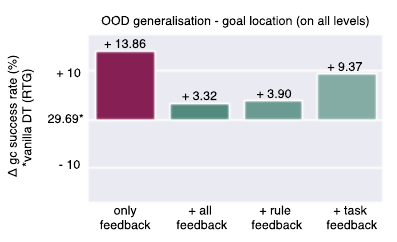}
\end{subfigure}%
\begin{subfigure}{.5\textwidth}
  \centering
\includegraphics{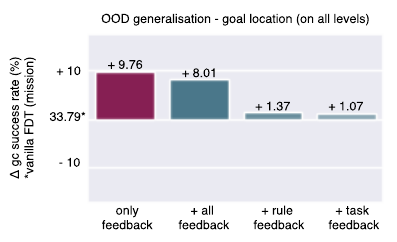}
\end{subfigure}
\caption{OOD generalisation performance for the proposed variants conditioning on return and/or feedback and mission and/or feedback compared against the return-only (vanilla DT) and mission-only (vanilla TDT) baselines, respectively, w.r.t. relative goal location across all levels.}
\end{figure}

\subsubsection{Impact of feedback at inference}

\begin{table}[htb]
    \centering
    \caption{Performance of the feedback-only variant against the mission-only and RTG-only baselines, and change in performance when using feedback at inference. *OOD performance averaged across OOD types.}
    \begin{tabularx}{\textwidth}{XXllll}
        \toprule
        \multirow{3}{*}{Baseline} & \multirow{3}{*}{Level} & \multicolumn{2}{l}{IID performance (\%)} & \multicolumn{2}{l}{OOD performance* (\%)}\\
        \cline{3-6}
        && Delta & Change & Delta & Change\\ 
        && (vs baseline) & (at inference) & (vs baseline) & (at inference)\\ 
        \midrule
        \multirow{9}{*}{mission} & All levels  & +4.96 & -2.34 & +4.96 & -4.99 \\ 
        & GoToObj  & +2.34 & -18.75 & -0.52 & -28.38 \\ 
        & GoToLocal  & +14.84 & -0.78 & +0.27 & +6.35 \\ 
        & PutNextLocal  & +4.88 & -0.39 & +9.62 & -6.00 \\ 
        & PickupLoc  & +1.95 & +10.94 & +5.20 & -2.47 \\ 
        & Pickup  & +8.99 & -5.47 & +9.57 & -0.39 \\ 
        & PutNext  & +4.88 & -2.35 & +6.45 & -4.11 \\ 
        & Synth  & -0.59 & -2.54 & +3.65 & -0.98 \\ 
        & SynthLoc  & +2.35 & +0.58 & +5.23 & -2.69 \\ 
        \midrule
        \multirow{9}{*}{RTG} & All levels & +0.05 & -2.34 & +2.33 & -4.99 \\ 
        & GoToObj & +3.91 & -18.75 & +1.56 & -28.38 \\ 
        & GoToLocal & +5.47 & -0.78 & -0.79 & +6.35 \\ 
        & PutNextLocal & +5.86 & -0.39 & +7.57 & -6.00 \\ 
        & PickupLoc & -8.98 & +10.94 & +1.95 & -2.47 \\ 
        & Pickup & +2.73 & -5.47 & +2.35 & -0.39 \\ 
        & PutNext & +1.17 & -2.35 & -0.39 & -4.11 \\ 
        & Synth & -0.98 & -2.54 & +2.28 & -0.98 \\ 
        & SynthLoc & -8.79 & +0.58 & +2.30 & -2.69 \\ 
        \bottomrule  
    \end{tabularx}
\end{table}

\begin{table}[htb]
    \centering
    \caption{Performance of the variants with all feedback in addition to mission/RTG against the respective baselines, and change in performance when using feedback at inference. *OOD performance averaged across OOD types.}
    \begin{tabularx}{\textwidth}{XXllll}
        \toprule
        \multirow{3}{*}{Baseline} & \multirow{3}{*}{Level} & \multicolumn{2}{l}{IID performance (\%)} & \multicolumn{2}{l}{OOD performance* (\%)}\\
        \cline{3-6}
        && Delta & Change & Delta & Change\\ 
        && (vs baseline) & (at inference) & (vs baseline) & (at inference)\\ 
        \midrule
        \multirow{9}{*}{mission} & All levels  & +4.28 & -7.20 & +2.41 & -6.63 \\ 
        & GoToObj  & +0.78 & -10.94 & -7.55 & -15.88 \\ 
        & GoToLocal  & +8.59 & -18.75 & -5.29 & -17.99 \\ 
        & PutNextLocal  & +4.88 & -4.10 & +6.15 & -0.63 \\ 
        & PickupLoc  & +2.73 & -3.51 & +5.86 & -6.90 \\ 
        & PutNext  & +6.25 & -3.52 & +5.66 & -3.58 \\ 
        & Pickup  & +6.65 & -11.33 & +11.33 & -7.04 \\ 
        & Synth  & -2.54 & +0.98 & +0.65 & -0.32 \\ 
        & SynthLoc  & +6.84 & -6.45 & +3.66 & -3.95 \\ 
        \midrule
        \multirow{9}{*}{RTG} & All levels & +2.64 & -1.15 & +2.59 & -4.07 \\ 
        & GoToObj & +25.78 & -9.37 & +19.53 & -14.32 \\ 
        & GoToLocal & +2.34 & +5.47 & +4.76 & -3.97 \\ 
        & PutNextLocal & +3.32 & -7.62 & +0.83 & -6.69 \\ 
        & PickupLoc & +1.18 & -0.78 & +2.87 & -5.73 \\ 
        & Pickup & +1.56 & -6.64 & +7.04 & -7.23 \\ 
        & PutNext & -2.74 & +0.20 & -3.65 & -1.04 \\ 
        & Synth & -1.37 & +5.47 & -3.91 & +4.10 \\ 
        & SynthLoc & -8.99 & +4.10 & -2.83 & +0.54 \\ 
        \bottomrule  
    \end{tabularx}
\end{table}

\begin{table}[htb]
    \centering
    \caption{Performance of the variants with Rule Feedback in addition to mission/RTG compared against the respective baselines, and change in performance when using feedback at inference. *OOD performance averaged across OOD types.}
    \begin{tabularx}{\textwidth}{XXllll}
        \toprule
        \multirow{3}{*}{Baseline} & \multirow{3}{*}{Level} & \multicolumn{2}{l}{IID performance (\%)} & \multicolumn{2}{l}{OOD performance* (\%)}\\
        \cline{3-6}
        && Delta & Change & Delta & Change\\ 
        && (vs baseline) & (at inference) & (vs baseline) & (at inference)\\ 
        \midrule
        \multirow{9}{*}{mission} & All levels  & +5.62 & -9.99 & +2.12 & -6.70 \\ 
        & GoToObj & -2.35 & -22.65 & -8.85 & -17.19 \\ 
        & GoToLocal & +4.68 & -8.59 & -2.11 & -17.99 \\ 
        & PutNextLocal & +7.22 & -6.25 & +6.45 & -3.32 \\ 
        & PickupLoc & +3.51 & -4.68 & +2.08 & +1.04 \\ 
        & Pickup & +10.16 & -10.16 & +5.47 & -0.39 \\ 
        & PutNext & +10.15 & -17.38 & +12.04 & -12.89 \\ 
        & Synth & +2.93 & -4.88 & +2.67 & -4.56 \\ 
        & SynthLoc & +8.60 & -5.28 & -0.39 & +0.20 \\ 
        \midrule
        \multirow{9}{*}{RTG} & All levels & +1.88 & -0.22 & +1.93 & -2.34 \\ 
        & GoToObj & +26.57 & -11.72 & +23.70 & -20.31 \\ 
        & GoToLocal & +8.60 & 0.00 & +2.91 & -1.06 \\ 
        & PutNextLocal & +0.78 & -4.88 & -4.00 & -2.05 \\ 
        & PickupLoc & -1.56 & -3.12 & -0.65 & -1.82 \\ 
        & Pickup & -3.91 & -3.13 & +2.35 & +1.37 \\ 
        & PutNext & -7.03 & +8.59 & -4.76 & +4.76 \\ 
        & Synth & -0.98 & +5.08 & -1.82 & +4.94 \\ 
        & SynthLoc & -7.43 & +7.43 & +0.34 & -3.12 \\ 
        \bottomrule  
    \end{tabularx}
\end{table}

\clearpage

\begin{table}[htb]
    \centering
    \caption{Performance of the variants with Task Feedback in addition to mission/RTG compared against the respective baselines, and change in performance when using feedback at inference. *OOD performance averaged across OOD types.}
    \begin{tabularx}{\textwidth}{XXllll}
        \toprule
        \multirow{3}{*}{Baseline} & \multirow{3}{*}{Level} & \multicolumn{2}{l}{IID performance (\%)} & \multicolumn{2}{l}{OOD performance* (\%)}\\
        \cline{3-6}
        && Delta & Change & Delta & Change\\ 
        && (vs baseline) & (at inference) & (vs baseline) & (at inference)\\ 
        \midrule
        \multirow{9}{*}{mission} & All levels  & -1.34 & +4.23 & -2.15 & +3.79 \\ 
        & GoToObj  & +2.34 & +3.13 & -9.37 & +10.41 \\ 
        & GoToLocal  & -13.29 & +18.76 & -24.07 & +21.96 \\ 
        & PutNextLocal  & -1.37 & +0.59 & +2.98 & -1.17 \\ 
        & PickupLoc  & -3.52 & +7.42 & +2.60 & +1.30 \\ 
        & PutNext  & -1.57 & +3.71 & -0.32 & +1.69 \\ 
        & Pickup  & +0.79 & +5.07 & +5.66 & +4.30 \\ 
        & Synth  & +1.37 & -1.76 & +2.15 & -2.93 \\ 
        & SynthLoc  & +4.50 & -3.13 & +2.25 & -1.42 \\ 
        \midrule
        \multirow{9}{*}{RTG} & All levels & +3.79 & -3.59 & +3.26 & -4.21 \\ 
        & GoToObj & +15.63 & -14.06 & +15.62 & -20.57 \\ 
        & GoToLocal & +7.03 & -1.56 & +5.29 & -3.17 \\ 
        & PutNextLocal & +3.12 & -5.85 & +1.27 & -6.74 \\ 
        & PickupLoc & +4.30 & +0.39 & +6.90 & +1.44 \\ 
        & Pickup & -3.13 & -1.17 & +0.79 & +2.73 \\ 
        & PutNext & +3.51 & -3.90 & +1.23 & -6.05 \\ 
        & Synth & +0.98 & -0.59 & +0.32 & -1.82 \\ 
        & SynthLoc & -1.18 & -1.95 & -3.32 & +1.71 \\ 
        \bottomrule  
    \end{tabularx}
\end{table}
\vfill

\end{document}